# Gradient Boosting on Decision Trees for Mortality Prediction in Transcatheter Aortic Valve Implantation

Marco Mamprin, Svitlana Zinger, Peter H.N. de With
Department of Electrical Engineering
Eindhoven University of Technology
Eindhoven, the Netherlands
m.mamprin@tue.nl,
s.zinger@tue.nl, p.h.n.de.with@tue.nl

Jo M. Zelis, Pim A.L. Tonino
Department of Cardiology
Catharina Hospital
Eindhoven, the Netherlands
jo.zelis@catharinaziekenhuis.nl,
pim.tonino@catharinaziekenhuis.nl

*Abstract*—Current prognostic risk scores in cardiac surgery are based on statistics and do not yet benefit from machine learning. Statistical predictors are not robust enough to correctly identify patients who would benefit from Transcatheter Aortic Valve Implantation (TAVI). This research aims to create a machine learning model to predict one-year mortality of a patient after TAVI. We adopt a modern gradient boosting on decision trees algorithm, specifically designed for categorical features. In combination with a recent technique for model interpretations, we developed a feature analysis and selection stage, enabling to identify the most important features for the prediction. We base our prediction model on the most relevant features, after interpreting and discussing the feature analysis results with clinical experts. We validated our model on 270 TAVI cases, reaching an AUC of 0.83. Our approach outperforms several widespread prognostic risk scores, such as logistic EuroSCORE II, the STS risk score and the TAVI2-score, which are broadly adopted by cardiologists worldwide.

## I. Introduction

Degenerative aortic valve stenosis (AS) is the most common valvular heart disease in the developed world. If left untreated the disease has a devastating course, rapidly causing death when symptoms develop. AS is caused by calcification of the aortic valve (AV). This could also lead to aortic valve regurgitation (AR) which also causes heart failure. The treatment for severe aortic valve disease consisted until recently of surgical aortic valve replacement (SAVR). However, in recent years TAVI has been developed and approved for use in severe to intermediate risk AV disease. Recently two randomized controlled trials have been published where use in low risk TAVI patients was non-inferior compared to SAVR. Despite the increasing development of this technique, there still is a risk bound to it. The broad use of TAVI in the last years has shown high chance of successful outcomes. However, the frailest patients sometimes do not benefit and can have complications after the procedure. The cause for this partial success is still not known.

So careful patient selection is paramount. Identifying those patients who have improvements or those who are at a higher risk after TAVI, is essential to maximize their survival, by providing an alternative treatment or therapy. Moreover, this would lead to an improvement in the use of the limited resources, which reduces the waiting lists. However, the identification of the patients that can have improvements from the TAVI procedure is a complex and still unsolved task because it is difficult to objectively quantify the improvements of a patient, in a daily routine. Furthermore, the patients that are at a higher risk, are often frail patients with several comorbidities and with an important medical history. Unfortunately, they can have severe complications, leading in the worst case to mortality, which further increases the motivation to find out why the identification was incorrect [1]. EuroSCORE II [2] and STS [3] risk scores are currently used for patient selection, but they were not developed for TAVI. Current predictors specifically designed for TAVI, such as TAVI2-SCORe [4] and Arnold SV et al. [5], do not offer optimal results. All these models are based on statistics.

With this paper, we aim to develop with a supervised machine learning approach, a very specific model to predict one-year mortality for the TAVI use-case. In fact, mortality at one year has been identified by the medical experts as the life expectancy threshold, above which the TAVI procedure is enabled and worth to be performed. We aim to validate its performance on a dataset of 270 patients that have undergone TAVI in 2015 and 2016.

We obtained promising prediction results and our work has the following contributions. Firstly, to the best of our knowledge, this is the first successful application of gradient boosting on decision trees for mortality risk prediction of TAVI. Secondly, by applying a state-of-the-art model explanation technique, we have created useful feature insight which allows the medical doctors and us to interpret the model and perform an optimal feature selection. Thirdly, we have validated our model on a retrospective clinical dataset and we have compared our prediction results with existing patient outcome predictors currently used in cardiac surgery.

## II. Methods

### A. Dataset

The anonymized dataset obtained from the Catharina Hospital consists of 270 TAVI procedures that were performed between January 2015 and December 2016. The average age of the patients, when TAVI procedure was performed, is 80.7 years, with a minimum age of 50.3 years and a maximum age of 94 years (standard deviation of 6.2 years). In our population, 48% of the patients are female.

The dataset was split into two categories and it has been classified a TAVI procedure as successful, when the patient survival was at least one year from the date of TAVI. Therefore, we split the patients into the following two classes:

− Survived within the first year (240 patients),
− Non-survived within the first year (30 patients, 50% of which did not survive within the first two months).

The dataset consists of numerous numerical and categorical information data, which can be divided in five principal categories: medical history, clinical data, patient questionnaire, risk score and medication.

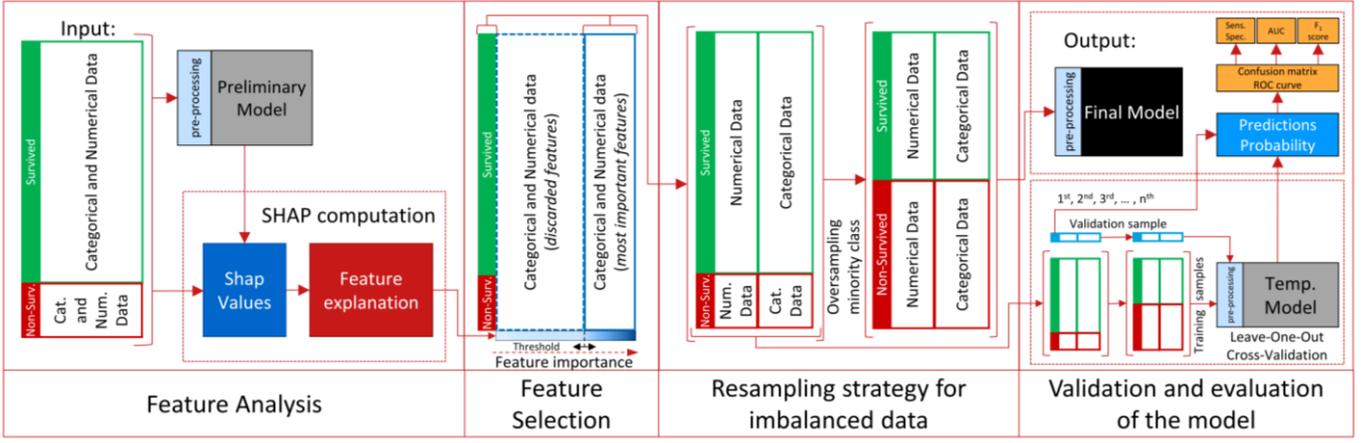

Fig. 1 Diagram of the one-year prediction model for the TAVI procedure (Implementation with Scikit-learn [16])

The retrospective study was approved by the local ethics committee and all enrolled patients signed informed consent.

*B. Initial processing of the dataset*

The overall processing steps of our data are depicted in Fig. 1. The initial processing described here is preceding the diagram and is based on discriminating between numerical and categorical features. Once each feature is defined, we analyze each individual feature in detail, by considering the mean, range and standard deviation of the numerical features, and the recurrence and instances of the categorical features.

We have reorganized some initially given features and some have been introduced, resulting from combining original features. For example, we mention the Body Mass Index (BMI) computed from the Height and the Weight. An alternative feature processing has been applied for multiple answer questionnaires by using one-hot encoding. An additional approach has been reserved for the list of prescribed medicines. After analysis with the medical experts, we added and divided them into 28 pharmacological macro-categories, according to their class of medication therapy and to their mechanism of functioning. Concerning the dates included in the original dataset, that were describing when certain events occurred prior to TAVI, it was decided to define them relative to the TAVI procedure date. Finally, we computed the intermediate and final scores for the RAND-36 questionnaire [6] according to the clinical guidelines.

*C. Gradient Boosting Algorithm*

Gradient Boosting on Decision Trees (GBDTs) is a supervised machine learning technique, which currently represents the state-of-the-art technique for models based on decision trees. Recently, a new categorical feature-specific model called CatBoost has been designed that outperforms several GBDTs algorithms [7], [8] such as: XGBoost (Chen and Guestrin, 2016) which does not have a dedicated pre-processing for categorical features and LightGBM (Keet *et al.*, 2017) which is not advisable on small-scale data.

The CatBoost model was used, since in clinical practice and consequently in this dataset, categorical features are common. Therefore, we can more easily exploit all the information provided in the dataset, leaving this innovative approach for pre-processing of the categorical and numerical features to the preliminary stage of CatBoost, see Fig. 1 at each input stage of each model. In fact, at the preliminary stage, CatBoost converts all the categorical features into numerical data by incorporating the recurrence of each instance, missing values (NaNs) are considered as an instance of the feature. Numerical features are then processed by aggregating different feature values in a histogram, which is specifically optimized for an efficient and fast memory access and elaboration. NaNs are processed as the minimum value of that feature to guarantee the split in the decision tree and to separate them more effectively from other numerical features.

The second stage of CatBoost involves an algorithm that builds an ensemble model with an iterative approach. At the $1^{st}$ iteration, the algorithm learns from the dataset the first decision tree, to reduce the training error. At the $2^{nd}$ iteration, the algorithm learns from the dataset one more decision tree, to reduce the error made by the decision tree obtained at the $1^{st}$ iteration and the algorithm repeats this procedure for all succeeding iterations until the iteration count is exceeded. This count is chosen to maximize the training of the model without overfitting the data and therefore reducing the generalization capabilities of the model.

*D. Feature Analysis and Selection*

The feature analysis stage is achieved by exploiting a new method that, with a local-level approach and its foundations in game theory, is able to provide interpretations and explanations of machine learning models, as shown in the first stage (left) of Fig. 1.

We exploited the SHapley Additive exPlanations unified approach [9], [10], using Shapley values, which is a technique recently published and applied to compute the importance of each feature of a model. Once extracted, all the importance values for each feature, we compute the following average:

$$\boldsymbol{\varphi}_m = \frac{\sum_{j=1}^{n}|\phi_{j,m}|}{n}, \quad (1)$$

where $\phi_{j,m}$ is the importance value of the feature $m$ for a patient $j$ on the total amount of $n$ patients. We obtain then the estimate of the most important features in the decision making of the model. Consequently, we reorder them from most relevant downwards by reordering the feature vector list.

The most important features were then discussed with the medical experts to find possible patterns and clinical explanations related to patient mortality. The validated relevant features which were confirmed by the clinical experts and that were found more discriminative in the decision process, have then been used to train the final model.

*E. Model training and resampling strategy*

At the feature analysis stage, we trained the model on a deep decision tree by including the entire dataset. We then analyzed the model to evaluate the importance of each feature. While the temporary and the final model has been trained only with the most informative features by imposing a threshold, to discard all non-relevant features. We conducted also a visual inspection of the SHapley Additive exPlanations summary plot, as shown in Fig. 3, which proved to be useful when discussing the results with the medical experts.

The training parameters of the final model have been chosen after performing a hyperparameter research, iterating different combinations of possible parameters and different levels of feature thresholds. We then identified which were more suitable for maximizing the $F_1$ score and AUC metric, by performing $k$-fold cross-validation multiple times with different randomizations per iteration. The temporary models used for validation utilize the same parameters as the final one.

The intrinsic imbalanced nature of the dataset, with a 1/8 ratio led us to apply a random over-sampling strategy to the minority class (non-survived patients) until a balanced ratio with the majority class (survived patients) was reached.

*F. Validation and Evaluation*

Validation of the model has been performed with 5 times Leave-One-Out Cross-Validation (LOOCV), with a dedicated pipeline for each iteration. In fact, the previously discussed random over-sampling of the minority class is applied at every iteration, but only after each test sample is removed.

Current state-of-the-art prognostic risk scores in cardiac surgery use the Area Under Curve (AUC) of the Receiver Operating Characteristic curve (ROC or C-statistic) as benchmark metric. However, this metric is not always the best choice, especially in the case of validation performed on a dataset with imbalanced classes. Therefore, we adopted other metrics such as sensitivity, specificity and $F_1$ score.

### III. EXPERIMENTAL RESULTS

*A. Selected Features*

According to the results obtained with the SHapley Additive exPlanations and the related discussion of these results with the cardiologists, we identified the most relevant features for the prediction of one-year mortality for TAVI.

The selected features are presented in alphabetical order and in order of importance, accordingly to the SHapley Additive exPlanations, in Table 1 and in Fig. 3, respectively.

Table 1 Details of the most important features for the prediction

| Description | Unit | Abbreviation | Mean value ± SD or Instances (sample size) | |
|---|---|---|---|---|
| | | | Survived (240) | Non-Survived (30) |
| AV regurgitation | - | ECHOAR | No (83) Mild (90) Moderate (19) Severe (13) | No (12) Mild (6) Moderate (6) Severe (6) |
| AV peak gradient | mmHg | AVPEAKGRAD | 75.21 ± 26.70 (171) | 64.87 ± 37.01 (21) |
| Atrioventricular block | - | AVBPRE | No (200) AV-block (34) AV-block III (2) | No (18) AV-block (7) AV-block III (2) |
| Beta blockers class of medicines | - | MEDBETAB | Yes (26) No (214) | Yes (11) No (19) |
| Body Mass Index | kg/m$^2$ | BMI | 26.93 ± 4.21 (239) | 26.17 ± 4.93 (30) |
| Creatinine | μmol/L | CREAT | 105.65 ± 50.14 (235) | 122.43 ± 51.30 (30) |
| General health score | 0-100 | QoLGENH | 37.67 ± 14.51 (104) | 22.14 ± 13.50 (7) |
| Haematocrit | % | HCT | 39 ± 5 (209) | 36 ± 3 (30) |
| Haemoglobin | mmol/L | HB | 7.89 ± 1.03 (205) | 7.38 ± 0.81 (29) |
| Month of post-procedure recovery | - | MONTH | Jan, …, Dec (16,21,20,21,14,28, 11,15,20,22,23,29) | Jan, …, Dec (3,3,2,2,2,1, 5,1,7,0,1,3) |
| Previous devices | - | PRVDEVICE | No (129) Yes (106) | No (7) Yes (4) |
| QRS duration | msec | TQRS | 109.67 ± 27.75 (234) | 122.81 ± 34.01 (26) |
| Smoking status | - | SMOKING | No (121) Former (45) Actual (14) | No (8) Former (3) Actual (1) |

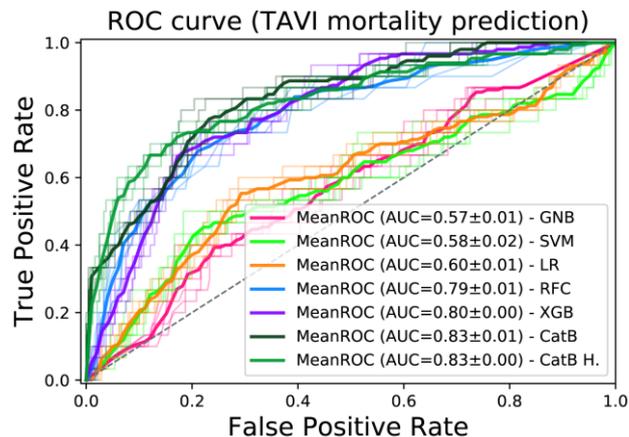

Fig. 2 ROC TAVI mortality prediction

Other relevant features that have been discarded to maximize the results at the risk of a reduced generalization capability on other datasets are: risk scores related to bleeding with antiplatelet and NSAIDs, NYHA class, QTC interval, gender, GFR, recent Coronary Artery Bypass Grafting (CABG) interventions, previous myocardial infarction and further class of medicines (e.g. ACE inhibitor, ARBs and PPIs).

We emphasize here that several factors are influencing the feature selection and the model training. Particularly, the order of the most important features can be highly influenced by the missing values. In fact, a certain feature can ascend or descend the list of the most important features proportionally to the amount of its missing values. It must be said that missing values contain information as well, because they are a natural consequence of the clinical workflow, of the internal protocols adopted by each hospital and of the varying urgency that characterize each patient journey. However, by including all the features shown in Table 1, in the final model, we ensured an optimal overview of the patient, acquired with a multi-disciplinary approach, which has shown promising generalization capability on this dataset.

*B. Evaluation of the model*

Current state-of-the-art prognostic risk score for the TAVI procedure is the TAVI2-SCORe [4] with an AUC of 0.72. EuroSCORE II and STS reached an AUC of 0.81 and 0.77, respectively. However, only 59 symptomatic patients with severe aortic stenosis were selected for this 30-day mortality study after the TAVI procedure [11].

We computed the ROC curve shown in Fig. 2 with an AUC of 0.83. We reached an accuracy of 0.90 with a specificity of 0.97 and sensitivity of 0.37. Finally, we computed an $F_1$ score of 0.45. All the evaluation metrics acquired during the validation of our prediction model are shown in Table 2 as value ± SD. As comparison, we re-performed the entire validation pipeline with other well-known machine learning algorithms. At this purpose, numerical values are scaled in the range ±1, replacing NaNs with 0 and One-Hot encoding is applied for categorical features, considering missing values as an instance apart.

Table 2 Results of the TAVI mortality prediction model based on CatB = CatBoost and comparison with XGB = XGBoost [12], RFC = Random Forest Classifier [13], LR = Logistic Regression, SVM = Support Vector Machine [14] and GNB = Gaussian Naive Bayes

| Metrics | CatB Hyper | CatB | XGB | RFC | LR | SVM | GNB |
|---|---|---|---|---|---|---|---|
| Sensitivity | 0.37 ± 0.06 | 0.33 ± 0.02 | 0.35 ± 0.04 | 0.14 ± 0.05 | 0.53 ± 0.01 | 0.49 ± 0.02 | 0.87 ± 0.00 |
| Specificity | 0.97 ± 0.00 | 0.98 ± 0.00 | 0.91 ± 0.01 | 0.97 ± 0.01 | 0.72 ± 0.01 | 0.74 ± 0.01 | 0.20 ± 0.01 |
| Accuracy | 0.90 ± 0.01 | 0.90 ± 0.00 | 0.85 ± 0.01 | 0.88 ± 0.01 | 0.70 ± 0.01 | 0.71 ± 0.01 | 0.27 ± 0.01 |
| $F_1$ score | 0.45 ± 0.06 | 0.44 ± 0.02 | 0.34 ± 0.03 | 0.21 ± 0.07 | 0.28 ± 0.01 | 0.27 ± 0.01 | 0.21 ± 0.00 |
| AUC-ROC | 0.83 ± 0.00 | 0.83 ± 0.01 | 0.80 ± 0.00 | 0.79 ± 0.01 | 0.60 ± 0.01 | 0.58 ± 0.02 | 0.57 ± 0.01 |

Table 3 Performance comparison of different mortality predictors

| Model name | Our approach | TAVI2-SCORe [4] | Arnold SV et al. [5] | EuroSCORE II [11] | STS Risk Score [11] |
|---|---|---|---|---|---|
| AUC | 0.83 | 0.72 | 0.66 | 0.81 | 0.77 |
| Population | 270 | 511 | 2830 | 59 | 59 |
| Mortality | one-year | one-year | one-year | 30-day | 30-day |

As shown in Table 3, our prediction model outperforms current state-of-the-art predictors, suggesting that prediction models based on machine learning can offer better performances. This especially holds for those difficult clinical cases where statistical approaches are not able to perform outstanding results. It should be noted that one-year mortality is generally more challenging than 30-day mortality.

## IV. DISCUSSION

### A. Clinical considerations

We evaluated each single feature with clinical experts, to provide explanations for each feature and the understanding of the possible interaction between all the different variables. With the goal of finding an answer to why certain features were considered more important than others in the decision making of the model, we infer the following conclusions.

(a) Severe *AV regurgitation* is linked to heart failure and mortality, which is found also in our analysis.
(b) A low *AV peak gradient* can be related to less severe aortic stenosis or to a dysfunctional left ventricle. High values are index of severe/critical aortic stenosis, that if untreated can lead to mortality for sudden cardiac death or heart failure.
(c) *Atrioventricular block* is linked to higher syncope and mortality and needs to be treated with a pacemaker giving complications, which reduces survival.
(d) *Beta blocker class of medication* are used to manage heart rhythms (atrial fibrillation or ventricular arrhythmia) and in case of heart failure. Both cases have an impact on survival.
(e) Extreme values of *Body Mass Index (BMI)* are unhealthy factors, even though there is clinical relevance that slight overweight gives a better prognosis after surgery.
(f) High blood *creatinine* concentration suggests that there is a kidney deficiency.
(g) Very low values for the RAND-36 [15] general health score indicates that the life quality of the patient is very low. This can be related to other comorbidities that reduce survival.
(h)(i) A low *hematocrit* and *hemoglobin* value may suggest anemia, which is linked to mortality.
(j) The *month* on which each patient faces its post-procedure recovery has shown to be statistically related to survival.

Certain cardiovascular complications are temperature related, while seasonal diseases also impact survival (comorbidities).
(k) *Previous devices* such as pacemaker, are implanted when rhythm disturbances are present and may lead to long-term complications. ICD or CRT-D are implanted when the left ventricular function is very low or when the patient has deadly arrythmia. Both these factors lead to a lower life expectancy.
(l) A high *QRS duration* is an index of a more diseased conduction system. This can lead to heart failure.
(m) *Smokers* have a higher risk of any cardiovascular disease, again a proof of the fact that they have less chance to survive after an operation.

## V. CONCLUSION

With this research, we aim to develop a predictive model for patient selection in cardiac surgery, specifically for the TAVI use-case, by exploiting a new state-of-the-art machine learning algorithm designed for categorical features. We have successfully identified the most important features to predict the one-year mortality and have developed a model based on the most relevant features. We validated our model on 270 patients, obtaining promising results and outperforming current predictive models based on statistics. The model enables to identify the patients that would have no chance of surviving to the first year with an AUC on the ROC of 0.83 and with a specificity and sensitivity of 0.97 and 0.37, respectively. Since our dataset was intrinsically imbalanced with a ratio of 1/8, we measured also the $F_1$ score as evaluation metric for this model, which resulted in a value of 0.45. It is important to extend the validation of the model on other patient data, possibly by performing a validation on a larger dataset, as well as in an inter-center validation process. Summarizing, we successfully developed a new prediction model based on machine learning for the TAVI procedure, offering attractive and powerful results for the prediction of one-year mortality. This forms a solid basis for further technical investigation and clinical studies.

ACKNOWLEDGMENT

This work is funded by EU ITEA, PARTNER project (No.16017).

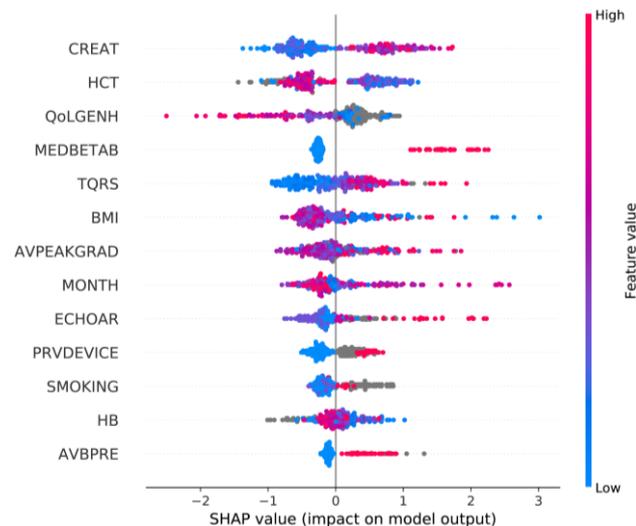

Fig. 3 SHapley Additive exPlanations summary plot (NaNs shown in grey)